\documentclass[letterpaper, 10 pt, conference]{ieeeconf}  

\IEEEoverridecommandlockouts                              

\usepackage{amsmath} 
\usepackage{subfigure,graphicx}
\usepackage{color}

\usepackage{multirow}
\usepackage{url}

\newcommand{\CameraReady}[1]{{#1}}
\title{\LARGE \bf
Beyond SIFT using Binary Features in Loop Closure Detection
}

\author{Lei Han$^1$, Guyue Zhou$^2$, Lan Xu$^1$, Lu Fang$^3$
\thanks{This work is supported in part by Natural Science Foundation of China (NSFC) under contract No. 61331015.}
\thanks{$^1$Lei Han and Lan Xu are with the Robotics Institute, Hong Kong University of Science and Technology, Clear Water Bay, Hong Kong {\tt\small \{lhanaf,lxuan\}@connect.ust.hk}}
\thanks{$^2$Guyue Zhou is with the DJI-Innovations, Shenzhen, China {\tt\small guyue.zhou@dji.com}}
\thanks{$^3$Lu Fang is with the Dept. of Automation, Tsinghua University, Beijing, China {\tt\small luvision.net@gmail.com}}
}

\begin{document}

\maketitle
\thispagestyle{empty}
\pagestyle{empty}

\begin{abstract}
In this paper a binary feature based Loop Closure Detection (LCD) method is proposed, which for the first time achieves higher precision-recall (PR) performance compared with state-of-the-art SIFT feature based approaches. The proposed system originates from our previous work Multi-Index hashing for Loop closure Detection (MILD), which employs Multi-Index Hashing (MIH)~\cite{greene1994multi} for Approximate Nearest Neighbor (ANN) search of binary features. As the accuracy of MILD is limited by repeating textures and inaccurate image similarity measurement, burstiness handling is introduced to solve this problem and achieves considerable accuracy improvement. Additionally, a comprehensive theoretical analysis on MIH used in MILD is conducted to further explore the potentials of hashing methods for ANN search of binary features from probabilistic perspective. This analysis provides more freedom on best parameter choosing in MIH for different application scenarios. Experiments on popular public datasets show that the proposed approach achieved the highest accuracy compared with state-of-the-art while running at 30Hz for databases containing thousands of images.
\end{abstract}

\section{INTRODUCTION}
\label{Intro}
Loop closure detection serves as a key component for globally consistent visual SLAM systems~\cite{mur2015orb}. Various approaches have been proposed to address this problem, but either suffer from low accuracy~\cite{galvez2012bags} and perform unstably under different scenarios, or from low efficiency~\cite{labbe2013appearance} and take too many computational resources to find a loop closure candidate. 

Binary feature (ORB~\cite{rublee2011orb} or BRISK~\cite{leutenegger2011brisk}) based methods~\cite{galvez2012bags,khan2015ibuild} benefit from their low computational complexity and efficient memory storage requirements, but also suffer from low precision and instability, showing a wide gap in accuracy compared with real-valued feature (SIFT~\cite{lowe2004distinctive} or SURF~\cite{bay2006SURF}) based approaches~\cite{labbe2013appearance,kejriwal2016high}. Our previously work MILD~\cite{han2017MILD} is the first binary feature based method that achieves comparable accuracy performance by employing MIH in ANN search for binary features instead of conventional methods that rely on a Bag-Of-Words (BOW) scheme. However, the accuracy of MILD is still inferior to state-of-the-art real-valued feature based approaches~\cite{labbe2013appearance},~\cite{kejriwal2016high}.

It is worthwhile to doubt \textbf{whether the ceiling of binary feature based LCD approaches has been reached} and {whether real-valued algorithms perform better than binary methods in terms of accuracy}. \cite{fan2016we} studied this problem in the context of 3D reconstruction, and argue that most of the popular binary features (ORB or BRISK) are inferior to SIFT based method in terms of reconstruction accuracy and completeness without a significant better computational performance.

We would argue that binary or real-valued features have different characteristics and should be treated differently. To exploit the potential of binary features in LCD approaches, the procedure of MILD is re-examined. A binary feature (ORB) based LCD approach that achieves better accuracy than SIFT based methods~\cite{labbe2013appearance,kejriwal2016high} is presented to support our idea.

\begin{figure}
\centering
\subfigure[The 199th image taken by left camera]{
\begin{minipage}[b]{0.22\textwidth}
\includegraphics[width=1\textwidth]{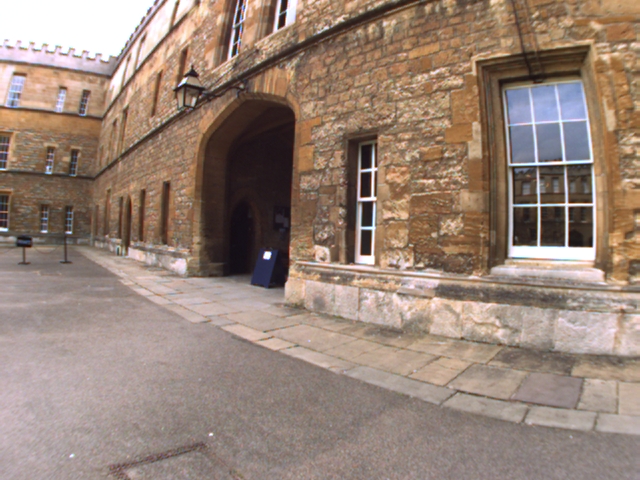} \\
\end{minipage}
}
\centering
\subfigure[The 199th image taken by right camera]{
\begin{minipage}[b]{0.22\textwidth}
\includegraphics[width=1\textwidth]{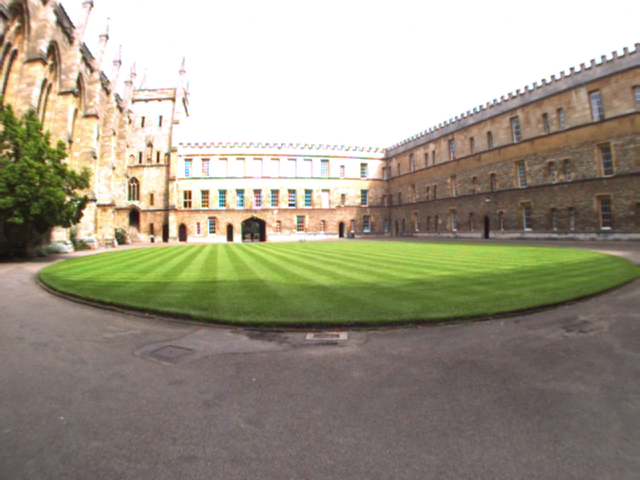} \\
\end{minipage}
}
\centering
\subfigure[The 300th image taken by left camera]{
\begin{minipage}[b]{0.22\textwidth}
\includegraphics[width=1\textwidth]{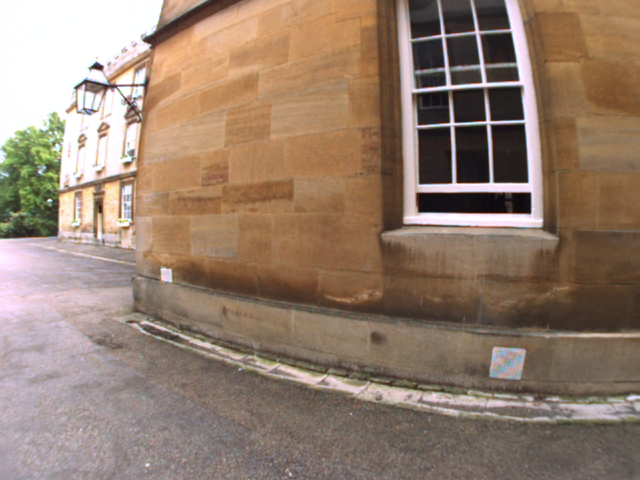} \\
\end{minipage}
}
\centering
\subfigure[The 300th image taken by right camera]{
\begin{minipage}[b]{0.22\textwidth}
\includegraphics[width=1\textwidth]{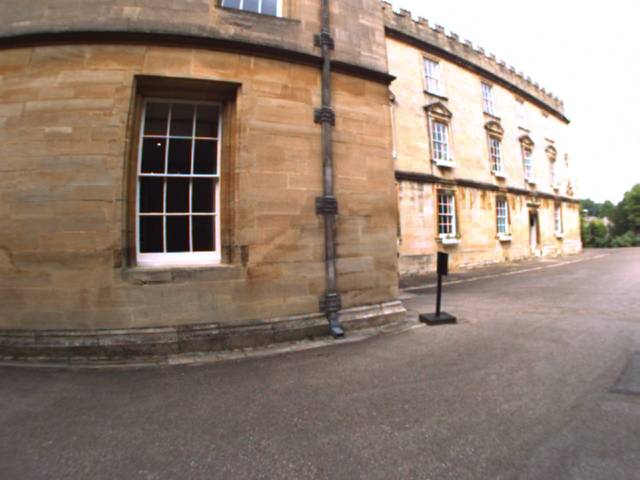} \\
\end{minipage}
}
\caption{Image~\cite{cummins2008fab} 199 and 300 are captured at different places, but shared a high similarity score.}
\label{fig:confused_image}
\end{figure}

One big challenge of MILD in achieving high accuracy is the inaccurate image similarity measurement. E.g., the two images shown in Fig.~\ref{fig:confused_image} are taken at different places while sharing a high similarity score. The same elements (windows, wall tiles) existing in different places tend to cause confusion and decrease the accuracy. Tracing to its source, we find that binary features are less discriminative compared with real-valued features since binary features carry much less information. The burstiness phenomenon~\cite{jegou2009burstiness} is more severe for binary features, which leads to a high similarity score for images taken at different places. Inspired by~\cite{jegou2009burstiness}, we propose burstiness handling procedure to overcome the misleading cases. No extra computational burden is introduced in this procedure by employing the specific features of MIH. 

While MIH is an important tool for ANN search of binary features in MILD, the analysis of MIH is incomplete as the search radius ($r$) in the substring hashing procedure is limited to $0$ for the lack of immediate probabilistic conclusions. As shown in~\cite{lv2007multi}, multi-probe techniques may lead to better performance (considering both accuracy and efficiency) for Locality Sensitive Hashing (LSH) of SIFT features. However, no theoretical analysis on the performance gain has been provided for the case of binary features. We propose a more comprehensive analysis of MIH by allowing arbitrary positive integers $r$ using the inclusion-exclusion-principle~\cite{ito1984introduction}. Based on this extension, the effect of Multi-Probe LSH for binary features is presented clearly as shown in Fig.~\ref{fig:recallComplexity}.

Even though MIH is very efficient in filtering out dissimilar features, a considerable part of the candidates selected using MIH would still be outliers for large datasets, a robust early termination strategy would reduce the frequency of memory access as well as Hamming distance calculation without loss of accuracy.

In summary, contributions of this paper include:

\begin{itemize}
\item
A more comprehensive theoretical analysis of MIH in ANN search of binary features is presented by enlarging the search radius $r$ in the substring hashing procedure. While in~\cite{han2017MILD}, $r$ is limited to zero. 
\item
The original MILD~\cite{han2017MILD} is improved in both accuracy and efficiency. Higher accuracy is achieved due to the dealing with burstiness of similar features, while the higher efficiency is due to the use of early termination and omit features selectively, based on probabilistic analysis. And, more remarkable, the designed burstiness handling technique is coped with the MIH procedure so that no extra computational burden is introduced in this procedure.
\item
Sparse match is proposed for frame matching using binary features, which is 16 times faster than traverse search at an average precision of 97\%, as shown in Sec.~\ref{sparse_match}.  
\end{itemize}

\section{RELATED WORKS}

\CameraReady{Various approaches have been proposed for loop closure detection, either using global image signatures~\cite{arandjelovic2013all} or local image features~\cite{angeli2008fast,cummins2008fab,galvez2012bags,khan2015ibuild}. While global signatures are relatively compact and can measure the similarity of two images efficiently, it is computationally expensive to extract global signatures, thus unpractical for on-line visual SLAM applications. Recently, deep-learnt features~\cite{cascianelli2017robust,hou2015convolutional} have also shown potentials in LCD for the ability of extracting semantic information from images. In this paper, we will focus on the local feature based approaches aiming to provide an efficient and robust loop closure detector for visual SLAM applications, as local features such as ORB~\cite{rublee2011orb} or SIFT~\cite{lowe2004distinctive} are widely used in visual SLAM systems for frame registration~\cite{mur2015orb,strasdat2012local} and they have been proven to be accurate and robust for various scenarios. }

Local feature based LCD approaches mainly explore on two sub-problems: feature classification and image similarity measurement. Feature classification tries to cluster different features indicating the same place, while image similarity measurement reveals the possibility of two frames indicating the same place based on previous observations. 
  
Most of the previous methods rely on the Bag-of-words (BOW) scheme for efficiency considerations. In BOW, features are clustered into different centroids (visual words) using offline/online trained dictionaries. Frames are represented by the histogram of visual words, and the loop closure likelihood is calculated based on the difference of visual word histograms. However, at least two well-known problems exist in such methods: 1. Perceptual Aliasing as features clustered into the same visual word may indicate different locations, and 2. They have high complexity for real-valued features while low accuracy for binary features. Among these methods, RTAB-MAP~\cite{labbe2013appearance} achieves the highest accuracy performance in terms of recall at 100\% precision, but takes $700$ ms per detection.~\cite{galvez2012bags} first tries binary features in the BOW scheme called bag of binary words, which is able to handle 20,000 images in $50$ ms with a much lower accuracy. 

Both MILD~\cite{han2017MILD} and~\cite{lynen2014placeless} argue the inefficiency of BOW for binary features. Unlike MILD using MIH~\cite{greene1994multi,norouzi2014fast} to replace BOW,~\cite{lynen2014placeless} simply project binary features into real-valued space which is feasible for the BOW scheme. However, the transformation from Hamming distance to Euclidean distance is highly nonlinear and a training procedure is required for choosing projection functions, which limits the range of applications for their proposed system.

Various methods have been proposed to improve the image similarity measurement including weak-geometric-check (WGC)~\cite{jegou2008hamming} and burstiness~\cite{jegou2009burstiness}.~\cite{jegou2008hamming} improves image retrieval quality by employing the rotation and scale information of local features.~\cite{jegou2009burstiness} handles the burstiness phenomenon of visual elements by assigning weights for features based on their similarity with others. However, such procedure requires a complex preprocessing to remove repeating features in each image. In this paper, we will show that combining the characteristics of MIH, burstiness can be handled with barely any computational burden.

\section{LOOP CLOSURE DETECTION}
\label{MILD}

As noted in~\cite{han2017MILD}, our previous work MILD is proposed for LCD by employing MIH in ANN search of binary features, which achieves significantly better PR performance compared with other binary feature based LCD approaches~\cite{galvez2012bags, khan2015ibuild}. However, the accuracy of MILD is still limited to repeating textures which is a common phenomenon in artificial or natural sceneries. The exploration of MIH for ANN search of binary features in~\cite{han2017MILD} is inadequate, as search radius $r$ in substring hashing procedure is limited to $0$.

For completeness, we will provide a brief review of MILD firstly. Then, based on the framework of MILD, the theoretical analysis of MIH for ANN search of binary features is further explored and extended in Sec.~\ref{multiprobeMIH} by considering multi-probe MIH. Next, a more precise model for image similarity measurement is provided in Sec.~\ref{burstinessHandling} to improve the accuracy of LCD. Finally, techniques on early termination are discussed in Sec.~\ref{earlyTermination} to improve the efficiency of MILD. An assumption throughout this paper is that the Hamming errors of two feature descriptors are evenly distributed, which will be used for the efficiency-accuracy analysis for MIH. 

\subsection{Review of MILD}
\label{reviewMILD}
In MILD, LCD is divided into two stages: image similarity measurement and Bayesian inference. Image similarity is used to compute the likelihood of two frames as a loop closure while Bayesian inference employs temporal coherency to get a final probability of loop closure based on image similarity. In this paper, we focus on improving the image similarity measurement, which is calculated by binary features directly and hence is more accurate than conventional methods using BOW representation of images. For a query image $I_p$, firstly binary features $F_p={f_1^p,f_2^p,\cdots,f_{|F_p |}^p}$ are extracted. The image similarity measurement between $I_p$ and $I_q$ is denoted as $\Phi(I_p,I_q )$:
\begin{align}
\Phi(I_p,I_q )=  \frac{\sum_{i=1}^{|F_p |}\sum_{j=1}^{|F_q|}\phi(f_i^p,f_j^q)}{|F_p ||F_q |},
\label{eqn:image_similarity_measurement}
\end{align}
where $\phi(f_i^p,f_j^q)$ refers to the binary feature similarity, i.e.,
\begin{align}
\phi(f_i^p, f_j^q) =
\begin{cases}
exp(-d^2/\sigma^2), d \leq d_0, \\
0,	d > d_0.
\end{cases}
\end{align}
here $d$ denotes the Hamming distance between binary features $f_i^p$ and  $f_j^q$, $\sigma$ is the weighting parameter, and $d_0$ is the pre-defined Hamming distance threshold. \CameraReady{$\Phi(I_p,I_q)$ evaluates the similarity of two images using a voting approach~\cite{jegou2008hamming}. The intuition behind Eqn.~(\ref{eqn:image_similarity_measurement}) is that if two images can be registered using binary features in visual SLAM, they tend to have feature pairs with small Hamming distance which results in a high similarity score.}

\begin{figure}[tbp]
\begin{minipage}[b]{1.0\linewidth}
  \centering
       \includegraphics[width=8cm]{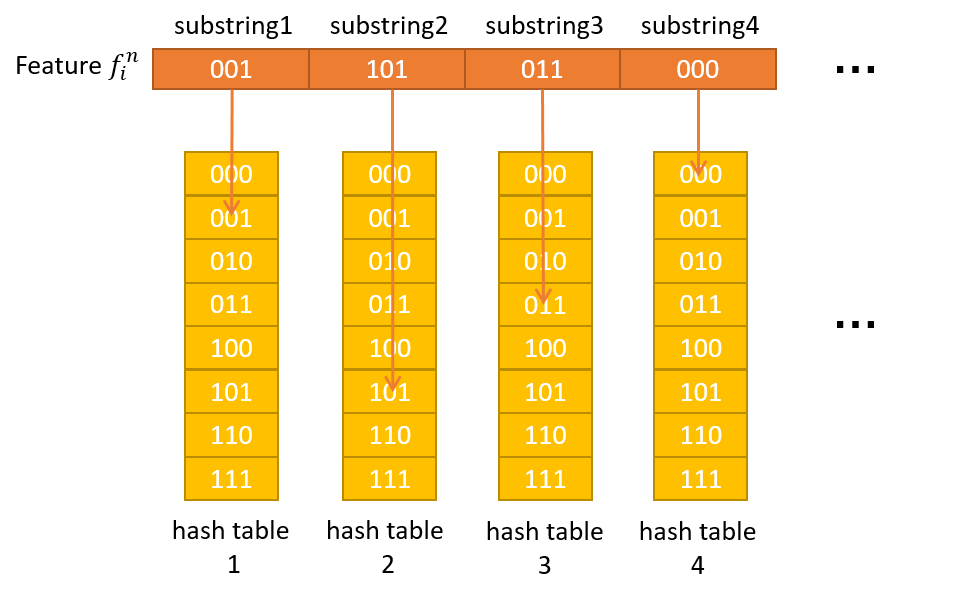}
\end{minipage}
\caption{Framework of MIH. Binary feature $f_i^n$ is divided into $m$ disjoint substrings. k-th substring is the hash index of the $k_{th}$ hash table. Image index $n$ and feature index $i$ are stored in corresponding $m$ entries as reference for feature $f^n_i$.}
\label{fig:MIH}
\end{figure}

Image similarities between $I_p$ and all candidate images stored in the database are computed using MIH, where a long binary feature descriptor $f_i^n$ is divided into $m$ disjoint substrings as shown in Fig.~\ref{fig:MIH}. Only features that fall into the same hash entry in at least $1$ hash table are considered as nearest neighbor candidates of $f_i^n$. Let $\Omega_i^n$ denote the collection of nearest neighbor candidates of $f_i^n$. Thus, image similarity measurement is approximated as:
\begin{equation}
\Phi'(I_n,I_k) = \frac{\sum_{i=1}^{|F_n|}\sum_{f_j^k \in \Omega_i^n}\phi(f_i^n,f_j^k)}{|F_n||F_k|}.
\label{Eqn:approximate}
\end{equation}

\cite{han2017MILD} evaluates the performance of MIH from two aspects: Complexity $E$ and Accuracy $R$. Complexity indicates the ratio between features counted in the candidate set and the total dataset. Accuracy is measured by the probability of two features representing the same location encountered in MIH. Lower $E$ leads to a more efficient image similarity computation while higher $R$ indicates the approximation is more accurate.

The probability that a feature pair with Hamming distance $d$ encountered in MIH is denoted as $P_{recall}$, which is a function of $d$, $m$ and $r$. $m$ is the number of hash tables and $r$ is the search radius in the substring hashing procedure. In~\cite{han2017MILD} only the situations of $r = 0$ are analyzed.
\begin{align}
P_{recall}(m,d) = 1 - \frac{m!\Theta(d,m)}{m^d}.
\end{align}

Prior statistics of Hamming distance distribution for ORB features~\cite{rublee2011orb} are employed in computing $E$ and $R$. For the ORB descriptor, the Hamming distance distribution for features of the same location (inliers) and different locations (outliers) are approximated as normal distribution $P_i(d) \in N(32,10)$ and $P_o(d) \in N(128,20)$, respectively. Thus, the complexity and accuracy of MIH can be computed as:

\begin{equation}
\begin{array}{cl}
R(m) &= \overset{L}{\underset{d=0}{\sum}} [P_{recall}(m, d)\times P_i(d)],	\\
E(m) &= \overset{L}{\underset{d=0}{\sum}} [P_{recall}(m, d)\times P_o(d)].
\end{array}
\label{Eqn:R_E}
\end{equation}

Best parameter of MIH ($m$) is chosen by considering both $R$ and $E$. Only features with the same substring of $f_i^n$ are selected into $\Omega_i^n$ in~\cite{han2017MILD}. The performance of MIH when features with similar substrings fall into the candidate set will be discussed in Sec.~\ref{multiprobeMIH}.

\subsection{Multi-Probe MIH}
\label{multiprobeMIH}

So far, only a few approaches have addressed the problem of fast search for binary features using hashing technique, including MIH~\cite{norouzi2014fast} for exact nearest neighbor search and LSH~\cite{muja2012fast} for ANN search. While in~\cite{han2017MILD} and this paper, the statistical information of MIH is employed for approximate nearest neighbor search. 

In LSH, multiple substrings are extracted from the original binary descriptor independently (e.g., binary elements are randomly selected from the original feature descriptor). To improve the search precision and maintain low complexity, multiple hash table and multi-probe hashing~\cite{lv2007multi} strategies are employed. Although extensive experiments conducted in~\cite{muja2012fast} show that the Hierarchical Clustering Tree (HCI) performs slightly better than LSH, HCI is not considered in our work. Because as a randomized algorithm, the performance of HCI may be unstable even on the same dataset. Beside this, for the problems of LCD, random selection of cluster centers is not possible as features are streamed into the database, it would be impractical to reorganize the tree structure every time a new feature enters.

MIH resembles LSH methods for it divides long binary codes into $m$ short substrings. Unlike LSH~\cite{muja2012fast} tries to make each substring independent from each other, MIH takes the inherent correlations in the substrings into consideration. For MIH, feature pairs with a Hamming distance less than $m$ will be discovered for sure.

Multi-probe Hashing~\cite{lv2007multi} has shown its ability to improve the search performance of LSH for real-valued features and is adopted in~\cite{muja2012fast} for fast match of binary features by enlarging the substring search radius $r$. However, in~\cite{han2017MILD}, only the situations where $r = 0$ are analyzed as no off-the-shelf mathematical tools can be found for $r > 0$. In this paper, we complete the analysis of MIH in ANN search of binary features by computing $P_{recall}(r,m,d)$ for arbitrary nonnegative integer $r$. The derivation is implemented in an iterative way, where the probability $P_{recall}(r, m, d)$ is computed from $P_{recall}(r-1,m,d)$ following the inclusion-exclusion-principle~\cite{ito1984introduction}. 
 
$P_{recall}(r, m, d)$ is equal to the probability that $d$ independent balls are thrown into $m$ bins randomly, \textbf{at least} one bin has \textbf{at most} $r$ balls. The situations of $r = 0$ have been analyzed in~\cite{han2017MILD}. Without loss of generality, $r=1$ equals to the union of two independent events:
\begin{enumerate}
\item{}
At least one bin has no ball.
\item{}
S bins have one ball in each bin, for the rest of the bins, each bin has at least one ball.
\end{enumerate}

The probability of event 1 ($P_1$) equals to $P_{recall}(0,m,d)$ computed in~\cite{han2017MILD}. Event $2$ is the union of $A_k, k = 1,2,\cdots,m$, where $A_k$ indicates that the $k_{th}$ bin has one ball, and the rest of the bins have at least one ball. Based on combinatorial analysis, we have
\begin{equation}
P(A_k) = \frac{d(m-1)^{d-1}(1-p_{recall}(0,m-1,d-1))}{m^d}.
\end{equation}

Following the inclusion-exclusion-principle, the probability of event 2 can be computed as:
\begin{equation}
\begin{aligned}
P_2 = &C_{N}^{1}P(A_1) - C_{N}^{2}P(A_1A_2) + C_{N}^{3}P(A_1A_2A_3) - \\
&\cdots  \pm C_{N}^{N}P(A_1A_2\cdots A_N).
\end{aligned}
\end{equation}

Then, $P_{recall}(1,m,d)$ is given by $P_1 + P_2$. Similarly, $P_{recall}(r,m,d)$ can be computed based on $P_{recall}(r-1,m,d)$ when r is larger than $1$. 

\begin{figure}[tbp]
\begin{minipage}[b]{1.0\linewidth}
  \centering
   \includegraphics[width=8cm]{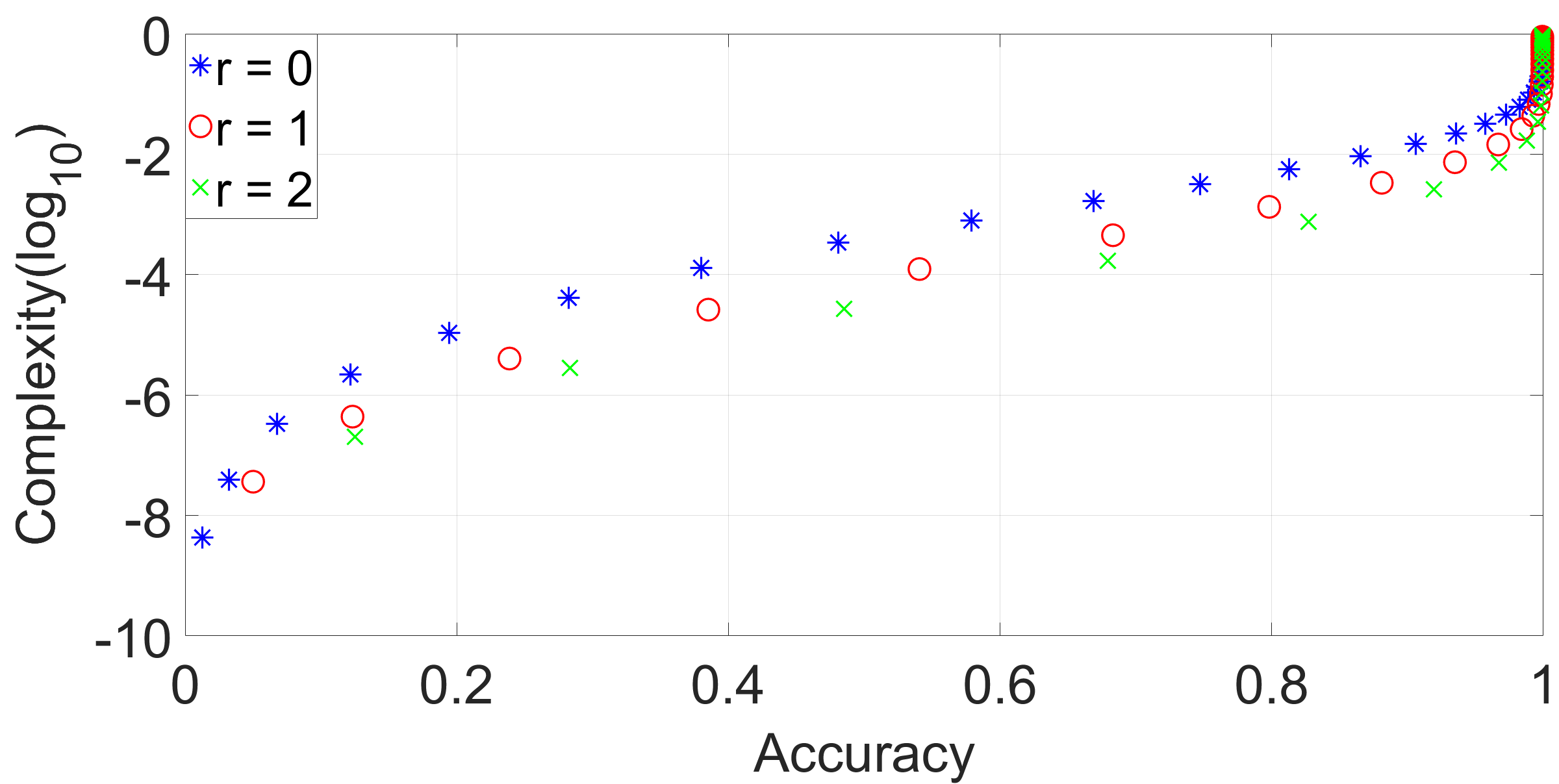}
\end{minipage}
\caption{Complexity and Accuracy under different MIH parameters including search radius $r$ and hashing table number $m$. Complexity and Accuracy grows monotonously with $m$ when $r$ is fixed. \CameraReady{$m$ starts with $4$ for different $r$ as shown in this figure.}}
\label{fig:recallComplexity}
\end{figure}

Given $P_{recall}(r, m, d)$, we can calculate the complexity and accuracy under different parameter configurations, $R(r,m)$ and $E(r,m)$, as shown in Fig.~\ref{fig:recallComplexity}. It can be concluded that:
\begin{itemize}
\item
A larger $m$ will lead to higher accuracy as well as higher complexity when $r$ is fixed;
\item
To achieve the same accuracy, it would be more efficient to use a smaller $m$ and a larger $r$.
\end{itemize}

Additionally, there is a fixed overhead for MIH: the construction of the hash tables, which is denoted by $E_{fixed}$. Suppose there are $m$ hash tables in MIH, each with $2^{\frac{L}{m}}$ hash entries ($L$ indicates the feature descriptor length, e.g., $L = 256$ for ORB features). $E_{fixed} = m2^{\frac{L}{m}}$ . 

From this analysis, given a required search precision, we can find the best parameter configurations for MIH to minimize the computational cost including both complexity $E$ and overhead $E_{fixed}$. E.g., in LCD problems, the minimum accuracy required is expected to be larger than $0.8$, three parameter candidates for $(m,r)$ include: $(15,0)$, $(10,1)$, $(8,2)$. For \textbf{on-line} loop closure detection problems, $(15,0)$ is more attractive as all the hash entries can be stored in memory. For applications when the hash tables can be built \textbf{off-line}, $(8,2)$ would be a better choice as the complexity of MIH is minimized and there is no need to consider $E_{fixed}$. In our implementations for LCD, $(16,0)$ is chosen as all the substrings can be represented as a $short$ data structure, which is more efficient for CPU operations.

\subsection{Burstiness Handling}
\label{burstinessHandling}
As discussed in~\cite{jegou2009burstiness}, visual elements may appear more times in an image than a statistically independent model would predict, which is described as the burstiness phenomenon. Burstiness corrupts the visual similarity measure in the context of image search. Two types of burstiness exist: intra-image, when repeating texture exists in the scenery; and inter-image, when similar objects exist in many places. Various weighting strategies were proposed to handle the burstiness phenomenon in~\cite{jegou2009burstiness}, while in this paper we combining the framework of MIH and handling burstiness without an increase of computational burden.

For intra-image burstiness, typically it may take a time-consuming preprocessing procedure to detect intra-image burstiness by a traversal of each feature extracted from the image~\cite{jegou2009burstiness}. Following the formulations in Sec.~\ref{reviewMILD}, the probability of independent features falling into the same hash entry is the complexity $E(r, m)$, which is approximated to $0$. While for repeating features, the probability is the accuracy $R(r, m)$, close to $100\%$ under our configurations ($r = 0, m = 16$). By limiting the number of features falling into the same hash entry for each image, we can accomplish the detection of intra-image repeating features without extra preprocessing procedure. 

For inter-image burstiness, we weight the feature similarity based on the total similarity score of the feature with all candidates provided by MIH. Given a query feature $f_i^n$, the feature similarity measurement is modified as
\begin{equation}
\phi'(f_i^n,f_j) = log(N/N_i)\frac{\phi(f_i^n,f_j)}{\sum_{f_k\in\Omega}\phi{f_i^n}{f_k}},
\end{equation}
where $log(\frac{N}{N_i})$ is the inverse document frequency term, $N$ represents the total number of candidate frames, and $N_i$ is the number of frames that have a similar feature with $f_i^n$.

Repeating features may cause inefficiency as these features may have little contribution to the image similarity measurement, but may exist in the candidate set frequently. Such inefficiency can be solved by limiting the maximum number of buckets $N_{buckets}$ in each hash entry. The average number of features falling into each hash entry would be: $A_{features} = \frac{N_{features}}{2^{l_s}}$, where $N_{features}$ is the maximum number of features stored in the dataset and $l_s$ is the substring length. $N_{buckets}$ is set to be $50 \times A_{features}$ by experiment. Entries that have more buckets than $N_{buckets}$ are discarded.

\subsection{Early Termination}
\label{earlyTermination}
The computational complexity of MILD grows linearly with the number of images stored in the database. For large datasets, the most time-consuming part of MILD is the memory access of feature descriptors and Hamming distance calculation. Given the sparsity of repeating locations, we have found that a majority of features in the candidate set are outliers with large Hamming distances. Based on this observation, we adopt early termination to avoid unnecessary memory access and computational cost for outliers by exploiting partial information of the feature descriptor. For the feature $f_i$ and its nearest neighbor candidate $f_j$, we only load the first $64$ bits of $f_j$ to calculate the partial Hamming distance $h_p$ between $f_i$ and $f_j$. If $h_p$ is larger than a threshold, $f_j$ is regarded as outliers directly instead of loading the rest bits of $f_j$. Experiments show that early termination succeeds to reject around $50\%$ of the outliers (Hamming distance is larger than $50$).

\section{EXPERIMENTS}
\label{experiments}

To evaluate the performance of our proposed approach, we conduct extensive experiments\footnote{Experiments are implemented on an Intel-core i7 @ 2.3 GHz processor with 8 GB RAM. Only one core is used to compare the computational efficiency of the proposed approach with other algorithms. For loop closure detection, up to 800 ORB features are extracted for each image using OpenCV. Feature descriptor length $L$ is set to be 256.} on different datasets\footnote{NewCollege~\cite{cummins2008fab} contains 1073 images of size $1280\times 480$. CityCentre~\cite{cummins2008fab} contains 1237 images of size $1280\times 480$. Lip6Indoor~\cite{angeli2008fast} has 388 images of size $192\times 240$. Lip6Outdoor~\cite{angeli2008fast} has 1063 images of size $192\times 240$. BovisaOutdoor~\cite{ceriani2009rawseeds} contains 2277 images of size $320\times 240$.}. Two individual experiments are implemented: LCD and frame match. Experiments in LCD reveal the superior performance of our method on loop closure detection. Experiments in frame match verify the precision of our proposed fast match algorithm: sparse match.

\begin{figure}[tbp]
\begin{minipage}[b]{1.0\linewidth}
  \centering
   \includegraphics[width=8cm]{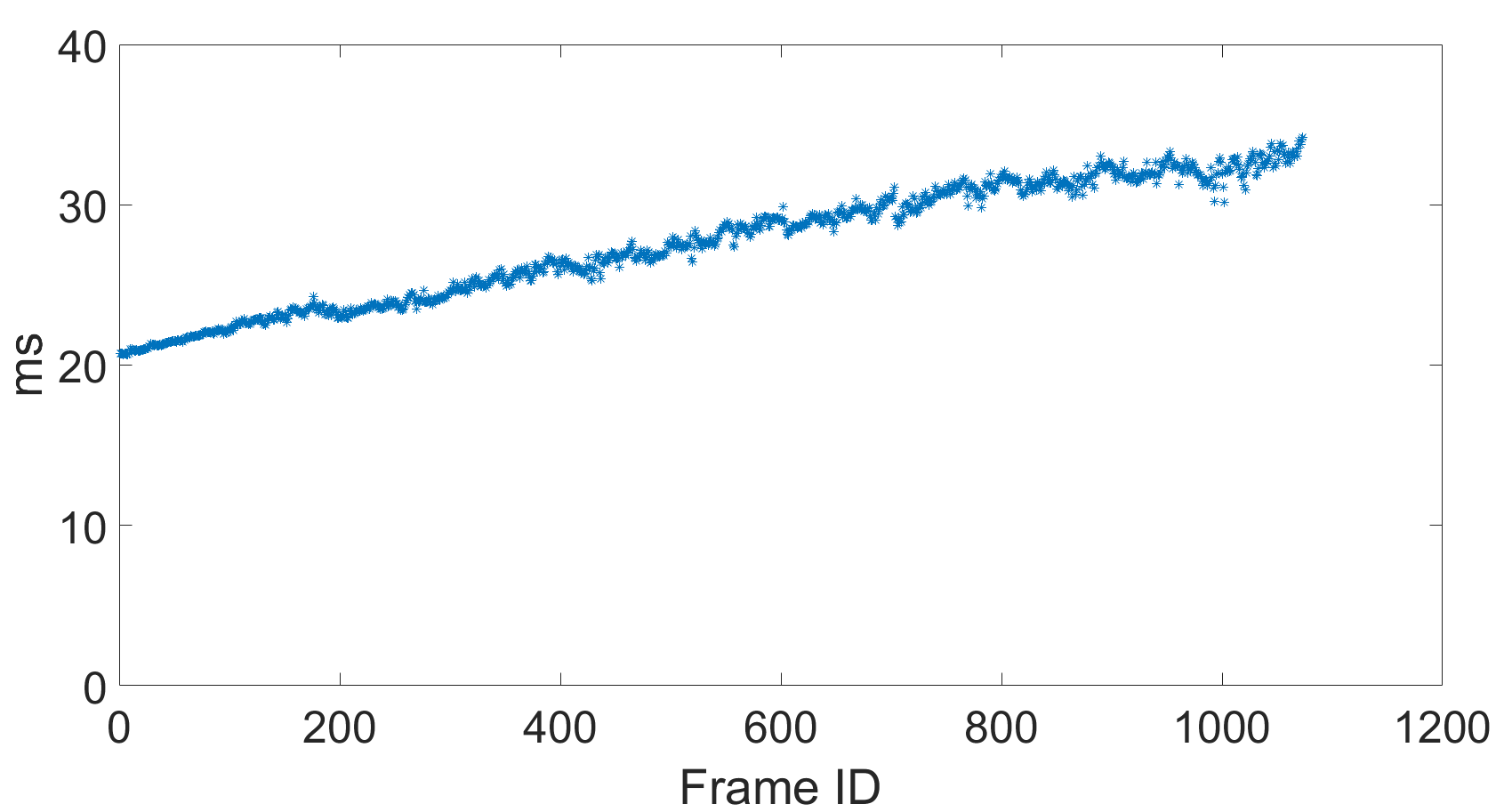}
\end{minipage}
\caption{\CameraReady{Computational complexity of the proposed method, in terms of running time (ms) for each frames for NewCollege dataset.}}
\label{fig:time_cost_per_frame}
\end{figure}
\subsection{Loop Closure Detection} 

\begin{figure}[tbp]
\begin{minipage}[b]{1.0\linewidth}
  \centering
   \includegraphics[width=8cm]{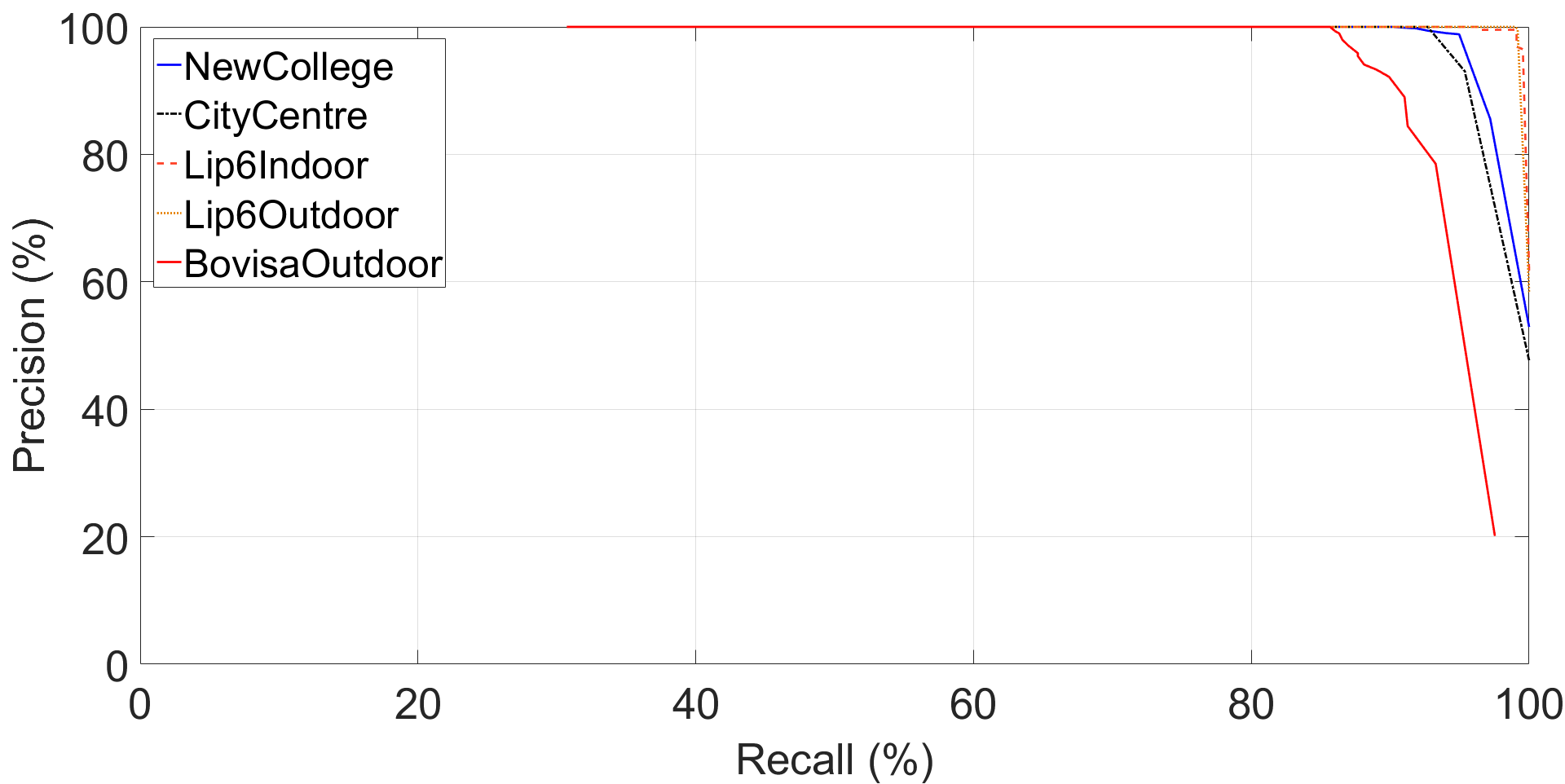}
\end{minipage}
\caption{Precision-recall curve for each dataset.}
\label{fig:precision_recall_curve}
\end{figure}

\begin{figure}[tbp]
\begin{minipage}[b]{1.0\linewidth}
  \centering
   \includegraphics[width=8cm]{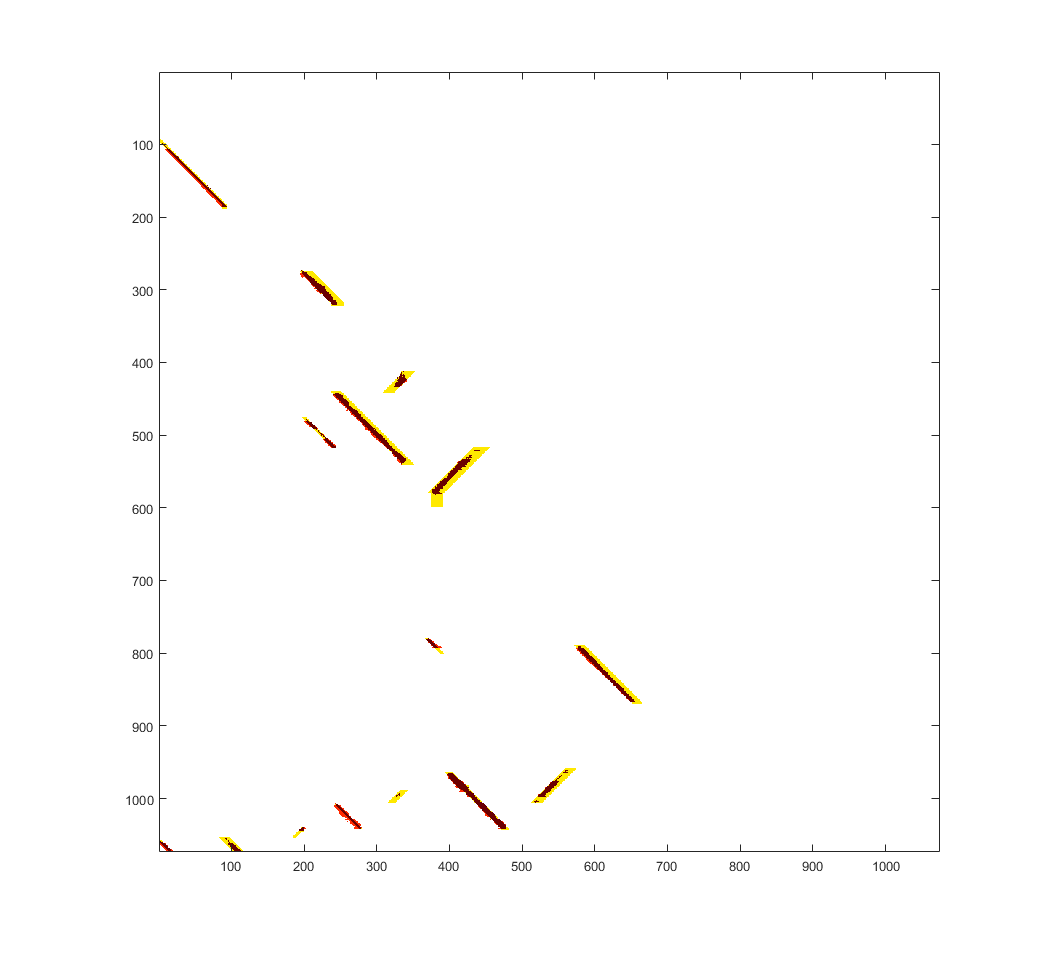}
\end{minipage}
\caption{Detected closure VS ground truth closure. The coordinate of each pixel $(i,j)$ represents the index for candidate image and query image respectively. Red indicates the detected closure by our approach, Yellow indicates the ground truth closure. Brown indicates the intersection of the detected closure and ground truth closure. All the red points are neighbors of the brown points and cannot be counted as false positives.}
\label{fig:DetectedLoops}
\end{figure}

Accuracy Evaluation: The precision-recall curves of all the datasets are provided in Fig.~\ref{fig:precision_recall_curve}. The proposed approach achieves high accuracy on all the datasets provided, including indoor, outdoor, natural and artificial sceneries.  In particular, we present the results on the NewCollege dataset to demonstrate the reliability of our proposed approach. As shown in Fig.~\ref{fig:DetectedLoops}. nearly all the ground truth closures are detected by the proposed method. 

Efficiency Evaluation: The runtime of LCD is composed by two main parts: $t_{f}$ for feature detection and extraction and $t_{q}$ for image similarity measurement. The other parts such as Bayesian inference can be completed efficiently within $1$ms, thus can be omitted. $t_{f}$ is fixed given image resolution and the number of features selected from each image, while $t_{q}$ grows linearly with the number of frames stored in the database. For the NewCollege dataset with $1073$ frames, the average runtime of each procedure is $t_f = 22.5$ ms and $t_q = 13$ ms for the original version of MILD. Based on the proposed early termination technique, $t_q$ can be reduced to $10$ ms without influencing of accuracy. The speed up factor would be even greater for memory IO inefficient systems, such as embedded chips or FPGA implementations. The time cost of LCD for each frame is also presented in Fig.~\ref{fig:time_cost_per_frame}. \CameraReady{Although the running time grows linearly with the number of candidate images, it can efficiently handle loop closures for datasets containing thousands of key frames as shown in the experiments, which is more than enough for visual SLAM designed for VR/AR applications or indoor navigation systems.}

To further verify the performance of proposed scheme, we compare our work with state of the art approaches, RTABMAP~\cite{labbe2013appearance} and BOWP~\cite{kejriwal2016high}, which are based on SIFT/SURF features, as well as DBOW~\cite{galvez2012bags}, IBuILD~\cite{khan2015ibuild} and MILD~\cite{han2017MILD} that use binary features, are used as references for comparison with the proposed system. The quantitative comparisons regarding accuracy (recall rate at $100\%$ precision) and runtime of the whole system (including feature extraction and loop closure detection) are shown in Table~\ref{tab:comparison}. Examining the results presented, the proposed approach achieves the highest recall rate on nearly all the datasets. The real-valued (SIFT or SURF) feature based approach RTABMAP~\cite{labbe2013appearance} ranks as the second place, while being 20 times slower than our proposed approach. \CameraReady{Note that for the BovisaOutdoor dataset, the proposed approach achieves much higher accuracy compared with the other approaches, because similar features (overexposure by the sun glare) exist in almost every frame, causing confusion in conventional LCD methods. Benefitting from the handling of burstiness, the repeating texture has negligible influence on the image similarity measurement.}

\begin{figure}[tbp]
\begin{minipage}[b]{1.0\linewidth}
  \centering
   \includegraphics[width=8cm]{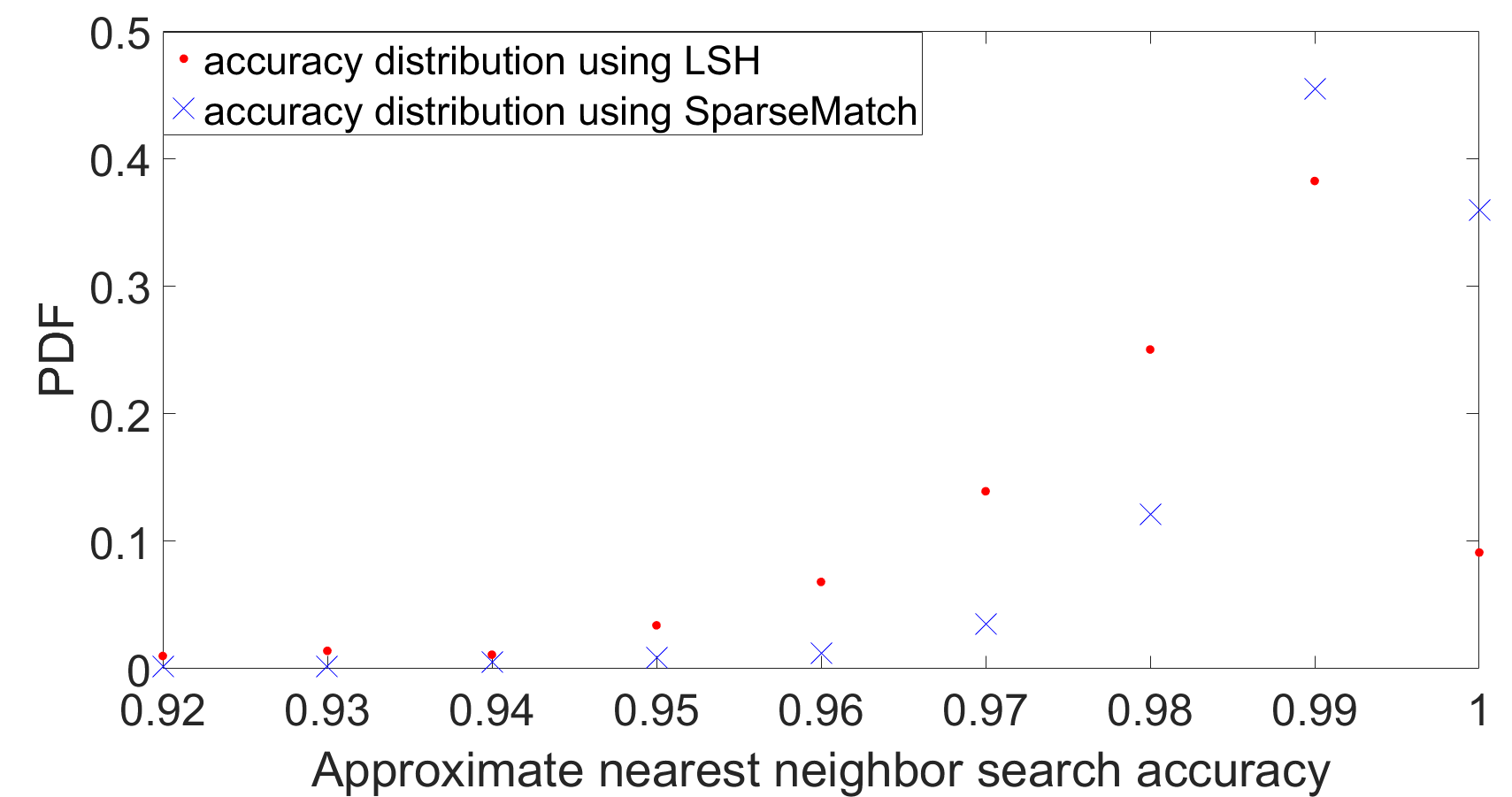}
\end{minipage}
\caption{Accuracy distribution of approximate nearest neighbour search using LSH and SparseMatch. Three nearest neighbour search methods are used in image matching: SparseMatch, LSH and brute-force method. Accuracy is evaluated by the percentage of correctly matched feature pairs in all valid feature pairs (Hamming distance less than 50).}
\label{fig:SparseMatch}
\end{figure}

\begin{table}[tbp]
\begin{center}
\label{tab:comparison}
\newcommand{\tabincell}[2]{\begin{tabular}{@{}#1@{}}#2\end{tabular}}
\begin{tabular}{|c|c|c|c|c|c|}
  \hline
  	& \tabincell{c}{City \\ Centre} & \tabincell{c}{New \\ College} & \tabincell{c}{Lip6 \\ Indoor} & \tabincell{c}{Lip6 \\ Outdoor} & \tabincell{c}{Bovisa \\ Outdoor} 
  \\
  \hline
  \multirow{2}{*}{RTABMAP~\cite{labbe2013appearance}} & 81\% & {89}\% & \textbf{98\%} & {95}\% & {52}\% \\ \cline{2-6}
  						  & 700ms & 700ms & 100ms & 400ms & 700ms
  \\
  \hline
  \multirow{2}{*}{BOWP~\cite{kejriwal2016high}} & {86}\% & 77\% & 92\% & 94\% & 40\% \\ \cline{2-6}
  						  & 441ms & 393ms & 69ms & 120ms &  209ms
  \\
  \hline
  \multirow{2}{*}{DBOW~\cite{galvez2012bags}} & 30.6\% & 55.9\% & - & - & 6\% \\ \cline{2-6}
  						  & 20ms & 20ms & - & - & 20ms
  \\
  \hline
  \multirow{2}{*}{CNN Feature~\cite{hou2015convolutional}} & 84.8\% & 82.4\% & - & - & - \\ \cline{2-6}
    						  & 155ms & 155ms & - & - & -
    \\
    \hline
  \multirow{2}{*}{IBuILD~\cite{khan2015ibuild}} & 38\% & - & 41.9\% & 25.5\% & - \\ \cline{2-6}
  						  & - & - & - & - & -
  \\
  \hline
  \multirow{2}{*}{MILD~\cite{han2017MILD}} & 83\% & 87.3\%  & 94.5\% & 93.4\% & 50\% \\ \cline{2-6}
  						  & 36ms & 35ms & 8ms & 9ms & 30ms
  \\
  \hline
  \multirow{2}{*}{Proposed} & \textbf{89.6\%} & \textbf{91.8\%}  & 97.8\% & \textbf{96.3\%} & \textbf{85.6\%}\\ \cline{2-6}
  						  & 33ms & 32ms & 8ms & 9ms & 25ms
  \\
  \hline

\end{tabular}
\end{center}\par
\bigskip
\CameraReady{Table 1.	Comparisons with state-of-the-art algorithms in terms of accuracy (recall rate at $100\%$ precision) and efficiency (average running time per-frame).}
\end{table}

\subsection{Frame Matching}
\label{sparse_match}
Benefit from our analysis on the efficiency-accuracy performance of MIH in ANN search of binary features, sparse match is proposed to find the corresponding features in images $I_1$ and $I_2$. For frame matching applications, overhead of the database construction must be small and high precision of the matching results is required. Conventional methods based on KD-trees~\cite{muja2012fast} are not suitable for such applications as they require a time-consuming overhead for initialization which can only work repeating usage situations. Sparse match is treated as a light version of LCD, where only $I_1$ is stored in the database, and $I_2$ is the query image. To minimize the overhead $E_{fixed}$ and maintain high precision $R$, the parameter of MIH used in sparse match is chosen as: $m = 24$, $r = 0$.

Experiments on $1000$ image pairs (consecutive images in the NewCollege dataset, $1200$ ORB features are extracted from each image) are implemented to verify the performance of MILD in terms of both accuracy and efficiency. In the experiments, three image match methods are implemented: sparse match, LSH implemented in FLANN and brute-force match. In LSH, we choose the same parameters as SparseMatch (24 hash tables, 10 bits for each key, multi-probe level equal to 0). As shown in Fig.~\ref{fig:SparseMatch}, SparseMatch achieved higher accuracy than LSH. The average processing time of each image pair for sparse match is $3.3$ ms including all overheads required, while $54.8$ ms for brute-force search and $14.88$ ms for LSH. The experiments are implemented on a lap-top computer and the number of available CPU cores is limited to 1 to compare the efficiency of different algorithms.

\section{CONCLUSIONS}
A binary feature based LCD approach is presented in this paper, which achieves the highest accuracy compared with state-of-the-art as shown in the experiments while running at 30Hz on a laptop. \CameraReady{Higher accuracy is achieved based on the handling of burstiness, which reduces the confusion caused by the repeating features. The complexity is also reduced by filtering outliers based on the partial Hamming distance.} The main bottleneck for efficiency of the proposed system lies in the extraction of binary features, which takes 70\% of the processing time. Fortunately, such computation can be shared with binary feature based SLAM systems. 

We will keep on improving the performance of the proposed LCD system and maintain the open-source implementation for the community\footnote{Code avaliable from: \url{https://github.com/lhanaf/MILD}}. \CameraReady{Currently, the proposed approach is only suitable for datasets containing thousands of keyframes. For larger datasets, the superiority of effiency may decrease as the complexity of MILD increases linearly with the number of candidates in the dataset. Memory management schemes such as~\cite{labbe2013appearance} can be further combined to enable the proposed algorithm running at constant time for large scale problems.} A better data structure that is suitable for large hash tables may also be adopted to further improve the efficiency of the proposed LCD system.

\bibliographystyle{IEEEbib}
\small{\bibliography{MILD2}}

\begin{thebibliography}{10}

\bibitem{greene1994multi}
Dan Greene, Michal Parnas, and Frances Yao,
\newblock ``Multi-index hashing for information retrieval,''
\newblock in {\em Foundations of Computer Science, 1994 Proceedings., 35th
  Annual Symposium on}. IEEE, 1994, pp. 722--731.

\bibitem{mur2015orb}
Raul Mur-Artal, Jose Maria~Martinez Montiel, and Juan~D Tardos,
\newblock ``Orb-slam: a versatile and accurate monocular slam system,''
\newblock {\em IEEE Transactions on Robotics}, vol. 31, no. 5, pp. 1147--1163,
  2015.

\bibitem{galvez2012bags}
Dorian G{\'a}lvez-L{\'o}pez and Juan~D Tardos,
\newblock ``Bags of binary words for fast place recognition in image
  sequences,''
\newblock {\em IEEE Transactions on Robotics}, vol. 28, no. 5, pp. 1188--1197,
  2012.

\bibitem{labbe2013appearance}
Mathieu Labbe and Francois Michaud,
\newblock ``Appearance-based loop closure detection for online large-scale and
  long-term operation,''
\newblock {\em IEEE Transactions on Robotics}, vol. 29, no. 3, pp. 734--745,
  2013.

\bibitem{rublee2011orb}
Ethan Rublee, Vincent Rabaud, Kurt Konolige, and Gary Bradski,
\newblock ``Orb: An efficient alternative to sift or surf,''
\newblock in {\em 2011 International conference on computer vision}. IEEE,
  2011, pp. 2564--2571.

\bibitem{leutenegger2011brisk}
Stefan Leutenegger, Margarita Chli, and Roland~Y Siegwart,
\newblock ``Brisk: Binary robust invariant scalable keypoints,''
\newblock in {\em 2011 International conference on computer vision}. IEEE,
  2011, pp. 2548--2555.

\bibitem{khan2015ibuild}
Sheraz Khan and Dirk Wollherr,
\newblock ``Ibuild: Incremental bag of binary words for appearance based loop
  closure detection,''
\newblock in {\em 2015 IEEE International Conference on Robotics and Automation
  (ICRA)}. IEEE, 2015, pp. 5441--5447.

\bibitem{lowe2004distinctive}
David~G Lowe,
\newblock ``Distinctive image features from scale-invariant keypoints,''
\newblock {\em International journal of computer vision}, vol. 60, no. 2, pp.
  91--110, 2004.

\bibitem{bay2006SURF}
Herbert Bay, Tinne Tuytelaars, and Luc Van~Gool,
\newblock ``Surf: Speeded up robust features,''
\newblock in {\em European conference on computer vision}. Springer, 2006, pp.
  404--417.

\bibitem{kejriwal2016high}
Nishant Kejriwal, Swagat Kumar, and Tomohiro Shibata,
\newblock ``High performance loop closure detection using bag of word pairs,''
\newblock {\em Robotics and Autonomous Systems}, vol. 77, pp. 55--65, 2016.

\bibitem{han2017MILD}
Lei Han and Lu~Fang,
\newblock ``Mild: Multi-index hashing for loop closure detection,''
\newblock {\em arXiv preprint arXiv:1702.08780, accepted in Multimedia and Expo
  (ICME), 2017 IEEE International Conference on}, 2017.

\bibitem{fan2016we}
Bin Fan, Qingqun Kong, Wei Sui, Zhiheng Wang, Xinchao Wang, Shiming Xiang,
  Chunhong Pan, and Pascal Fua,
\newblock ``Do we need binary features for 3d reconstruction?,''
\newblock in {\em Proceedings of the IEEE Conference on Computer Vision and
  Pattern Recognition Workshops}, 2016, pp. 53--62.

\bibitem{jegou2009burstiness}
Herv{\'e} J{\'e}gou, Matthijs Douze, and Cordelia Schmid,
\newblock ``On the burstiness of visual elements,''
\newblock in {\em Computer Vision and Pattern Recognition, 2009. CVPR 2009.
  IEEE Conference on}. IEEE, 2009, pp. 1169--1176.

\bibitem{cummins2008fab}
Mark Cummins and Paul Newman,
\newblock ``Fab-map: Probabilistic localization and mapping in the space of
  appearance,''
\newblock {\em The International Journal of Robotics Research}, vol. 27, no. 6,
  pp. 647--665, 2008.

\bibitem{lv2007multi}
Qin Lv, William Josephson, Zhe Wang, Moses Charikar, and Kai Li,
\newblock ``Multi-probe lsh: efficient indexing for high-dimensional similarity
  search,''
\newblock in {\em Proceedings of the 33rd international conference on Very
  large data bases}. VLDB Endowment, 2007, pp. 950--961.

\bibitem{ito1984introduction}
Kiyosi It{\^o},
\newblock {\em An Introduction to Probability Theory},
\newblock Cambridge University Press, 1984.

\bibitem{arandjelovic2013all}
Relja Arandjelovic and Andrew Zisserman,
\newblock ``All about vlad,''
\newblock in {\em Proceedings of the IEEE conference on Computer Vision and
  Pattern Recognition}, 2013, pp. 1578--1585.

\bibitem{angeli2008fast}
Adrien Angeli, David Filliat, St{\'e}phane Doncieux, and Jean-Arcady Meyer,
\newblock ``Fast and incremental method for loop-closure detection using bags
  of visual words,''
\newblock {\em IEEE Transactions on Robotics}, vol. 24, no. 5, pp. 1027--1037,
  2008.

\bibitem{cascianelli2017robust}
Silvia Cascianelli, Gabriele Costante, Enrico Bellocchio, Paolo Valigi, Mario~L
  Fravolini, and Thomas~A Ciarfuglia,
\newblock ``Robust visual semi-semantic loop closure detection by a
  covisibility graph and cnn features,''
\newblock {\em Robotics and Autonomous Systems}, vol. 92, pp. 53--65, 2017.

\bibitem{hou2015convolutional}
Yi~Hou, Hong Zhang, and Shilin Zhou,
\newblock ``Convolutional neural network-based image representation for visual
  loop closure detection,''
\newblock in {\em Information and Automation, 2015 IEEE International
  Conference on}. IEEE, 2015, pp. 2238--2245.

\bibitem{strasdat2012local}
Hauke Strasdat,
\newblock {\em Local accuracy and global consistency for efficient visual
  slam},
\newblock Ph.D. thesis, Citeseer, 2012.

\bibitem{lynen2014placeless}
Simon Lynen, Michael Bosse, Paul Furgale, and Roland Siegwart,
\newblock ``Placeless place-recognition,''
\newblock in {\em 2014 2nd International Conference on 3D Vision}. IEEE, 2014,
  vol.~1, pp. 303--310.

\bibitem{norouzi2014fast}
Mohammad Norouzi, Ali Punjani, and David~J Fleet,
\newblock ``Fast exact search in hamming space with multi-index hashing,''
\newblock {\em IEEE transactions on pattern analysis and machine intelligence},
  vol. 36, no. 6, pp. 1107--1119, 2014.

\bibitem{jegou2008hamming}
Herve Jegou, Matthijs Douze, and Cordelia Schmid,
\newblock ``Hamming embedding and weak geometric consistency for large scale
  image search,''
\newblock in {\em European conference on computer vision}. Springer, 2008, pp.
  304--317.

\bibitem{muja2012fast}
Marius Muja and David~G Lowe,
\newblock ``Fast matching of binary features,''
\newblock in {\em Computer and Robot Vision (CRV), 2012 Ninth Conference on}.
  IEEE, 2012, pp. 404--410.

\bibitem{ceriani2009rawseeds}
Simone Ceriani, Giulio Fontana, Alessandro Giusti, Daniele Marzorati, Matteo
  Matteucci, Davide Migliore, Davide Rizzi, Domenico~G Sorrenti, and Pierluigi
  Taddei,
\newblock ``Rawseeds ground truth collection systems for indoor
  self-localization and mapping,''
\newblock {\em Autonomous Robots}, vol. 27, no. 4, pp. 353, 2009.

\end{thebibliography}

\end{document}